\documentclass[twoside,11pt]{article}

\usepackage[abbrvbib, preprint]{jmlr2e}

\usepackage{jmlr2e}
\usepackage[table,xcdraw]{xcolor}
\usepackage{amsfonts}
\usepackage{amsmath} 

\jmlrheading{}{}{}{03/15/2020}{}{}{Jung Hoon Lee}

\ShortHeadings{Library network for explainable networks }{Lee}
\firstpageno{1}

\begin{document}

\title{Library network, a possible path to explainable neural networks}

\author{\name Jung Hoon Lee \email jungl@alleninstitute.org \\
       \addr Allen Institute for Brain Science\\
       Seattle, WA 98109, USA
       }

\editor{}

\maketitle

\begin{abstract}
Deep neural networks (DNNs) may outperform human brains in complex tasks, but the lack of transparency in their decision-making processes makes us question whether we could fully trust DNNs with high stakes problems. As DNNs’ operations rely on a massive number of both parallel and sequential linear/nonlinear computations, predicting their mistakes is nearly impossible. Also, a line of studies suggests that DNNs can be easily deceived by adversarial attacks, indicating that their decisions can easily be corrupted by unexpected factors. Such vulnerability must be overcome if we intend to take advantage of DNNs’ efficiency in high stakes problems. Here, we propose an algorithm that can help us better understand DNNs' decision-making processes. Our empirical evaluations suggest that this algorithm can effectively trace DNNs' decision processes from one layer to another and detect adversarial attacks.
\end{abstract}


\section{Introduction}

Deep neural networks (DNNs) trained via deep learning (DL) have been adopted in a growing number of domains; see \citep{Lecun2015} for a review. DNNs are considered highly efficient in solving problems because they do not require any detailed instructions from users and can outperform humans in some domains. For instance, DL was successfully used to train `AlphaGo’ to learn a complicated board game `Go’ and defeat Sedol Lee, one of the best Go players \citep{wiki:alphago}. This match demonstrated DNNs/DL’s efficiency, but it also revealed that their operations/decision-making processes are not transparent. Both professional Go players and AlphaGo creators still do not comprehend AlphaGo’s adopted strategies against Sedol Lee. 

Notably, efficiency alone cannot justify deploying DNNs into all domains. In high stakes problems, safety is more important than efficiency, and it remains unclear whether DNNs’ operations could warrant safety \citep{Lipton2016,Rudin2018}. First, DNNs rely on a massive number of parallel and serial numerical operations, making diagnosis of their failures impossible. Second, DNNs’ decisions can be easily corrupted by adversarial attacks \citep{Akhtar2018,Goodfellow2015,Huang2017}. Therefore, before we deploy DNNs into high stakes problems demanding rigorous decision-making, we need to place safety measures. To this end, it is imperative that we better understand how they reach their decisions. 

A line of studies proposed that properties of hidden neurons, representing intermediate stages of DNNs’ decisions, provide insights into their operating principles and decision-making processes. First, feature visualization was proposed to study hidden neurons’ response characteristics \citep{olah2017feature,carter2019activation}. This approach allows users to identify optimal visual features that can stimulate target hidden neurons, which will advance our understanding of DNNs' learning processes; see also \citep{Erhan2009, Simonyan2013}. Second, features of hidden layers were proposed to study DNNs’ operations \citep{Alain2016,Montavon2011,Liu2018}. The layer-specific features were analyzed by clustering algorithms and linear classifiers. For instance, Allain and Bengio \citep{Alain2016} tested linear-separability of features in hidden layers and found that linear separability increases monotonically, when the selected layer is closer to the last layer.  Third, a set of single neurons’ responses, evoked by multiple examples, was used as feature vectors. The responses were different from those of multiple neurons evoked by the same input. Raghu et al. \citep{Raghu2017}  used them to study how strongly the hidden layers are correlated with ground truth (i.e., the labels of inputs).

As hidden neurons represent intermediate states of decisions, it seems natural to assume that they encode crucial information in the neural codes underlying the decisions and that these codes can help us build more explainable and safer DNNs. Then, how do we find such codes? To address this question, we propose an algorithm that can predict DNNs’ answers on unseen examples (i.e., test examples) from hidden layer activity patterns (HAPs). Our motivation is as follows: if an algorithm can predict DNNs' answers, it must capture the crucial codes for their decisions. Our empirical evaluations showed that our newly proposed algorithm can reliably predict DNN’s answers, supporting that it can identify the neural codes crucial for their decisions. Furthermore, we find that this new algorithm can also be used to detect adversarial attacks.

\section{Methods}

We assume that 1) HAPs’ meanings remain relative and become apparent only when HAPs are compared with one another and 2) that HAPs evoked by the same-class inputs are clustered together; importantly, the number of HAPs' clusters is not necessarily equal to the number of classes. If these assumptions are correct, proper and effective characterization of HAPs can shed light on internal mechanisms of DNNs.  Here, to effectively catalogue HAPs, we turn to our earlier short-term memory systems \citep{Lee2018}. 

Our short-term memory systems \citep{Lee2018} consist of an input, synaptic and output layers, and they store novel input patterns (i.e., substantially different from stored input patterns) by imprinting them into synaptic weights. Specifically, the inputs are normalized so that the output nodes of short-term memory can estimate the cosine similarity between the present and stored input patterns. Consequently, when all outputs of short-term memory are substantially lower than 1, the present input pattern is novel one. Otherwise, the present input pattern is similar to previously presented input patterns. 

In this study, we use this short-term memory systems, referred to as `library networks' hereafter, to inspect HAPs' diversity and study their links to DNNs' decisions. Below, we explain how library networks are constructed and how we use them to explain DNN's answers.

\subsection{Network structure}\label{fs}
\subsubsection{ Generating libraries of activity patterns in hidden layers}
\begin{figure}[ht]
	
	\begin{center}
		\centerline{\includegraphics[width=5.5in]{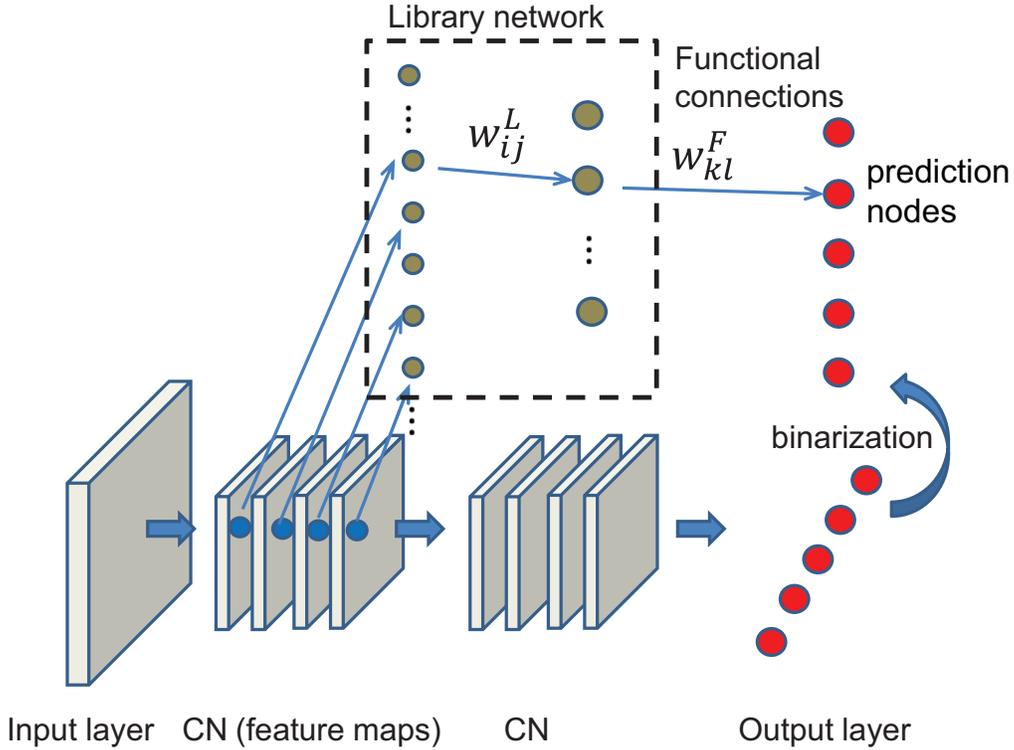}}
		\caption{The structure of library networks. The top part of the figure shows the schematics of a library network, whereas the bottom part shows a DNN consisting of multiple convolutional layers (CNs) and a fully-connected (FC) layer (chosen for illustration). For clarification, a single library network is displayed. CN layers produce multiple sheets of feature maps, and all nodes within them are mapped onto input nodes of the library network (indicated by a dashed box). Similarly, all outputs of FC layers are introduced to the library networks. The output nodes of the library networks are determined by the properties of input patterns (outputs of hidden neurons). Then, the outputs of the library networks are correlated with DNNs' outputs (shown in red circles). In the experiments, we also construct the library networks for blocks of layers used in ResNet. The outputs of these blocks are modulated with batch normalization before introduction to the library networks.}
		\label{Fig1}
	\end{center}
	
\end{figure}

As shown in Fig. \ref{Fig1}, a library network has a single synaptic weight layer and accepts normalized HAPs (Eq. \ref{eq1}) as inputs. 
\begin{equation}\label{eq1}
h^k_{i}=\sum_{j}w^L_{ij}\frac{f^k_j}{\left\Vert\vec{f^k}\right\Vert}, \text{where } w^L_{mn}=\frac{f^m_n}{\left\Vert\vec{f^m}\right\Vert}
\end{equation}
, where $h^k_{i}$ represents a synaptic input to output node $i$ of library networks evoked by the input pattern  $\vec{f^k}$.

It performs two main tasks. First, it estimates synaptic input ($h^k_i$) evoked by input patterns and finds maximum values among them. When maximal activation values are below the predefined threshold value ($\theta$), the inputs are labeled as novel. Second, it stores novel inputs by adding output nodes and imprinting inputs to synaptic weights targeting newly added nodes (Eq. \ref{eq1}); that is, a newly added node effectively `stores' a present input pattern. As output nodes are continuously added, its size is determined by characteristics of HAPs and the threshold value; naturally, the higher the chosen threshold value $\theta$ ($\leq 1$) is, the bigger the constructed library networks are.

In our experiments, we construct library networks for multiple hidden layers while introducing all training examples. Once the library networks are fully constructed with the training set, they can estimate cosine similarities between a present HAP and stored ones. If the previously stored HAP is presented again to the library networks, the input $h^k_i$ to node $i$, added when presented, will be 1, but all other activation functions are smaller than 1. 

\subsubsection{Functional connections between library networks and decisions}\label{s2.2}
The motivation of using library networks can be easily explained with an extreme case, in which HAPs' clusters are well separated according to classes (i.e., desired labels) of input patterns. If we further assume that HAPs in the same clusters are extremely similar to one another but significantly different from those in other clusters, we can naturally assume that a single (representative) HAP can represent a class of input pattern. That is, library networks will store one HAP pattern for each class. After all representative HAPs are stored in library networks, the number of library network's output nodes will be the same as the number of classes of input patterns. More importantly, any input pattern will evoke a HAP similar to one of representative HAPs. Thus, library networks' outputs can be used to infer the class of the input pattern. 

In DNNs, such an extreme case may not hold, but we assume that training can enforce HAPs to cluster together depending on classes of input patterns, and thus the library networks can still predict the classes of inputs. If this hypothesis is valid, we can use library networks' predictions to evaluate how well DNNs are trained. Furthermore, we can gain insights into functions of individual layers of DNNs and their contributions to DNNs' decisions-making processes.  

To address this possibility, we construct (linear) prediction nodes, which make predictions on input patterns' classes based on the outputs of library network. In this study, to evaluate the utility of library networks in explaining DNNs' decision-making, we train prediction nodes to predict DNNs' answers by establishing the Hebbian connections (Eq. \ref{eq2}) between the prediction nodes and the output nodes of library networks.

\begin{equation}\label{eq2}
W^F_{mn}=\sum_k g(h^k_{n})\times O^k_m, \text{where } g(x)=\exp\left( \frac{-(1-x)}{0.01} \right) 
\end{equation}
, where $W^F_{mn}$ denotes the functional connection from the library network node $n$ to prediction node $m$; where $O^k_m$ and $h^k_n$ represent the input to prediction node $m$ (depending on DNNs' answers) and the input to output node $n$ of the library network when $k$th input pattern is presented. During the construction of these Hebbian connections (that can capture correlations between library networks' outputs  and DNNs' answers), the input of prediction node corresponding to the DNN's answer (i.e., predicted class of an input pattern) is 1, and the rest of the inputs are -1; that is, if the DNN predicts that an input pattern $k$ belongs to a class $c$, $O^k_c$=1 and $O^k_{j \neq c}=-1$. It should be noted that $W^F_{mn}$ depends on nonlinear kernels (Eq. \ref{eq2}) to render sharper correlations. 

Specifically, after constructing library networks, we feed training examples to DNNs and evaluate both HAPs and DNNs' answers. For each training example, we update connections  $W^F_{mn}$ according to Eq. \ref{eq2}. This training is completed after a single iteration; no example is used twice for the training of $W^F_{mn}$. Consequently, the learning of prediction nodes is fast. 

With these functional connections, we test the library networks' capability to predict DNNs' decisions on unseen (training) examples. Specifically, we  calculate the likelihood ($P^k_m$) of the answer being $m$ in response to $k$th input by using three or eight maximally activated output nodes of the library networks (Eq. \ref{eq3}) 

\begin{equation}\label{eq3}
P^k_m=\sum^{a=3,8}_{a=1} W^F_{ma}\times \text{sort} (h^k)_a 
\end{equation}   
, where sort$(h^k)_a=(h^k_{argmax(h^k)[0]},h^k_{argmax(h^k)[1]},...)$; $argmax(V)$ represents indices of vector $V$ components sorted in descending order. 

\subsection{Empirical evaluations of library networks' utility} \label{ss}
In this study, we test the library networks' utility with two DNNs, 1) a convolutional network (CNN) trained with MNIST dataset \citep{LeCun1998} and 2) a ResNet trained with CIFAR10 dataset \citep{Krizhevsky2009}. Both CNN and ResNet are implemented using `Pytorch', the open-source python machine learning libraries \citep{Paszke2017}. CNN used here is an implementation of a variation of LeNet-5 \citep{LeCun1998}, and we adopt the official implementation of pytorch \citep{Pytorch-team2018} and train it with default parameters from the released example. For the ResNet, we use pre-trained networks of ResNet44 available in the public github repository \citep{Idelbayev2018}; ResNet44 is referred to as ResNet. 

MNIST includes 60,000 training and 10,000 test images of handwritten digits (0–9). Each image consists of 28-by-28 8-bit gray pixels. CIFAR-10 is the collection of 10 classes of items ranging from animals to man-made objects. Each class has 500 training and 100 test examples, each of which is a 32-by-32 color image. MNIST and CIFAR-10 can be obtained from http://yann.lecun.com/exdb/mnist/ and  https://www.cs.toronto.edu/~kriz/cifar.html, respectively. Using these datasets and networks, we evaluate the performance of the library networks, constructed for CNN and ResNet, when predicting their answers. All codes used in this study are freely available in the public github repository \citep{Lee2019}. 

\section{Results}
\subsection{Peeking into CNN through library networks}
\begin{figure}[ht]
	
	\begin{center}
		\centerline{\includegraphics[width=5.5in]{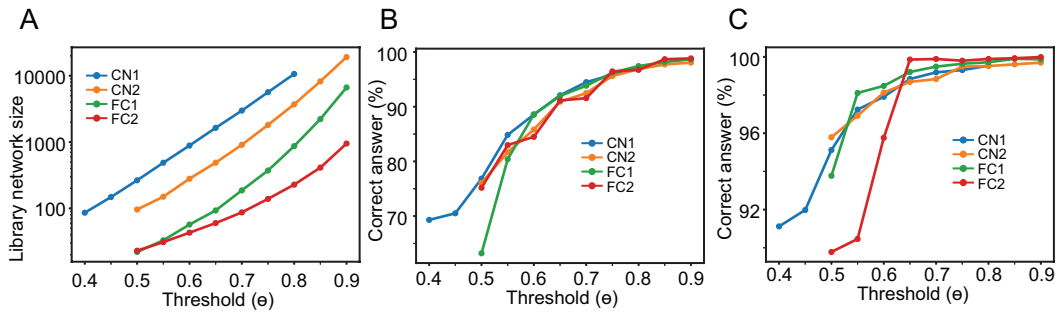}}
		\caption{Empirical evaluations using the CNN trained for MNIST. (A), The sizes of the library networks (i.e., the number of output nodes) depending on threshold values for novelty detection (see the text). The color codes are used to specify target layers. For instance, CN1 represents a library network's size using HAPs from the first CN layer. (B), The fraction of correct predictions of the library networks using single best answers. (C), The same as (B) but the three best answers are used for predictions.}
		\label{Fig2}
	\end{center}
	
\end{figure}

CNN used here consists of 2 convolutional (CN) layers and 2 fully-connected (FC) layers. As the size of the library networks reflects HAPs' homogeneity, we first measure the sizes of 4 library networks for CN1, CN2, FC1 and FC2 layers in the CNN, depending on threshold values ($\theta$). When the same threshold value is used for all layers, the library networks of the earlier hidden layers are bigger than those of the later layers (Fig. \ref{Fig2}A); for instance, the library network for FC1 is bigger than the others when a threshold value is chosen for all layers. That is, HAPs in the earlier layers are more heterogeneous, suggesting that the input vectors are transformed into homogeneous ones (possibly, homogeneous clusters), as information propagates through CNN. 

Next, using the Hebbian rule (Eq. \ref{eq2}), we train prediction nodes for all 4 library networks with CNN's answers on training examples, rather than the true labels of examples. Once the prediction nodes are trained, we test how well the library networks can predict DNNs' answers/decisions (Section \ref{s2.2}). Specifically, to estimate $P^k_m$ in Eq. \ref{eq3}, we use 3 maximally activated output nodes of the library networks for CNN. Prediction is based on the identity of maximally activated prediction node. For instance, if $P^5_1$ is the biggest among $P^5_n$, where $n=0,...,9$, the 5th input is predicted to be the digit `1'. Fig. \ref{Fig2}B shows the prediction accuracy from all 4 library networks for CN1, CN2, FC1 and FC2 depending on $\theta$. As shown in the figure, the library networks can reliably predict CNN's answers on test examples, especially when the higher threshold values are chosen. We further test the predictive power of the library networks by allowing them to provide three best answers (the three digits corresponding to three biggest $P^k_m$). If CNN's decision coincides with one of the three best answers proposed by the library networks, we count it as a correct answer. As shown in Fig. \ref{Fig2}C, the library networks exhibit dramatically enhanced predictive power. 

We note that the predictive power of the library networks for the later layers are higher than those for the earlier layers. This is consistent with the notion that input patterns are transformed progressively to networks' decisions through layers (see Alain and Bengio \citep{Alain2016}, for instance). As the library networks seem to represent information regarding CNN's decisions, we further track changes in predictions from one layer to another by establishing the confusion index ($CI$) between the two digits using the inputs to prediction nodes in each layer. $CI(d_1,d_2)$ is formally defined in (Eq. \ref{eq4}). By definition, $CI(d_1,d_2)$ evaluates the ratio of the likelihood of the answer being $d_2$ to the likelihood of the answer being $d_1$. 

\begin{equation}\label{eq4}
CI(d_1,d_2)=\langle \exp(P^k_{d_2})/ \exp(P^k_{d_1}) \rangle_k, 
\end{equation}
where $P^k_i$ represents the input to $i$th prediction node in response to $k$th input pattern (eq. \ref{eq3}); for instance, the first and last prediction nodes represent digits 0 and 9, respectively; where $\langle A \rangle_k$ represents average of $A$ over input patterns $k$.

In our experiments, we use trials (i.e., forward passes) to estimate $CI$, in which a maximally activated prediction node (Eq. \ref{eq3}) correctly predicts the digit ($d_1$) presented to the CNN. As such, $CI$ estimates the probability of the library networks' misidentifying the digit $d_1$ as $d_2$, on average. As $CI(d_1,d_1)$ is always 1, we do not report them below. 

\begin{figure}[ht]
	
	\begin{center}
		\centerline{\includegraphics[width=5.5in]{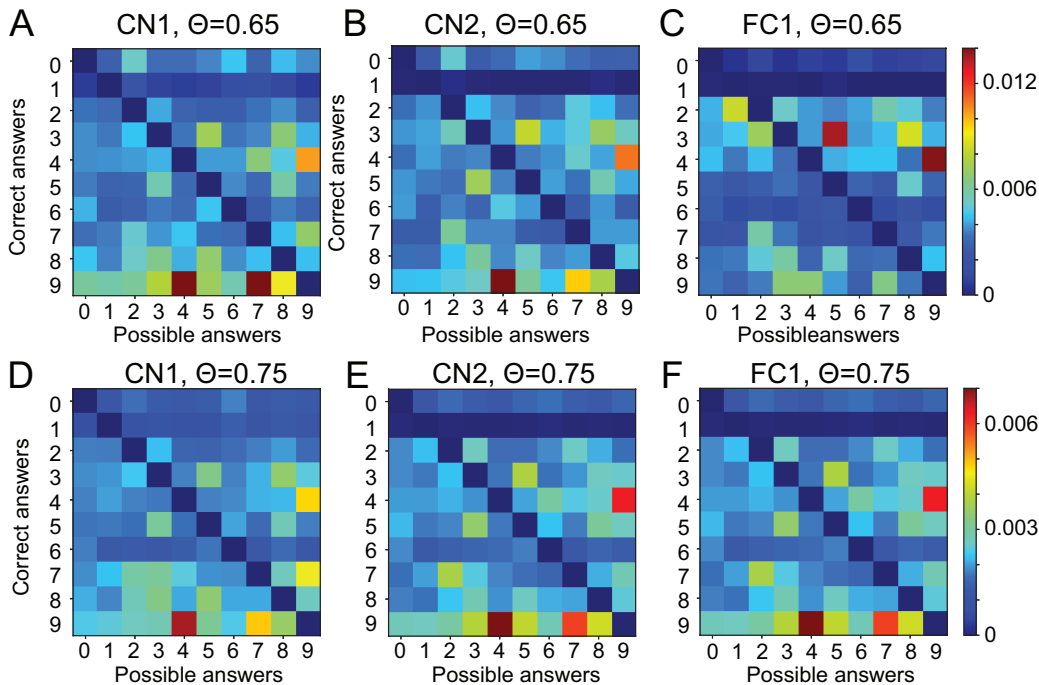}}
		\caption{Confusion matrix. (A)-(C), The confusion indices estimated using the library networks' predictions constructed for CN1, CN2, FC1, respectively. $y$-axis represents the correct digit ($d_1$), and $x$-axis represents a possible answer ($d_2$). For instance, the values shown in the 10th row and 5th column represent the probability of the library networks' reporting digit `4' in response to digit `9'. The colors represent $CI (d1,d2)$, from 0 to 0.014. (D)-(F), The same as (A)-(C), but the threshold value is 0.75 instead of 0.65.  The colors represent $CI (d1,d2)$, from 0 to 0.007.}
		\label{Fig3}
	\end{center}
\end{figure}

Figure \ref{Fig3} A-C show the confusion matrices whose components are $CIs$, when $\theta=0.65$. We do not display $CI$ for FC2 because they are extremely low. The y-axis represents $d_1$ (correct answer), and the x-axis, $d_2$, respectively. As shown in the figure, digit 9 confuses CN1 layer (i.e., the first convolution layer). Interestingly, CN2 also shows relatively high $CI$ values when digit 9 is presented (i.e., $d_1=9$). These confusions are alleviated in FC1 layer. To determine whether this trend is a consequence of low $\theta$, we increase $\theta$  from 0.65 to 0.75 and find equivalent results (Fig. \ref{Fig3} D-F). The high $CI$ values in response to digit 9 ($d_1=9$) can be explained by the fact that digit 9 has multiple visual features similar to those in other digits. Thus, CN layers, which rely on filters with limited spatial dimensions, may have difficulty differentiating 9 from others. FC layers, fully connected to previous layers, can use global features (e.g., locations of features) to precisely recognize digits. We also note that $CI(4,9)$ is high in all layers, which seems natural given the similarity between 4 and 9.  These results support the possibility that CN layers are optimized to identify local features, while FC layers are optimized to utilize the (relative) locations of the features detected by CN layers. Thus, we propose that the library networks and confusion matrices help us infer the workflow of DNNs consisting of CN and FC layers.

\subsection{Peeking into ResNet through library networks}
We further test the library networks' utility using a ResNet trained with CIFAR 10 dataset. Specifically, we use the pretrained ResNet44 network from the public github repository (section\ref{ss}). Although there are 44 layers in the ResNet, it is organized with 5 functional blocks. The three blocks, composite layers (CL) 1, 2 and 3, include multiple layers, and the rest (CN1 and FC) are single layers. The library networks are constructed to inspect the 5 blocks instead of all hidden layers, and we repeat the same experiment. In the ResNet, we used the 8 highest output activation values of the library networks to calculate the input to prediction nodes $P^m_k$ (Eq. \ref{eq3}). 

\begin{figure}[ht]
	
	\begin{center}
		\centerline{\includegraphics[width=5.5in]{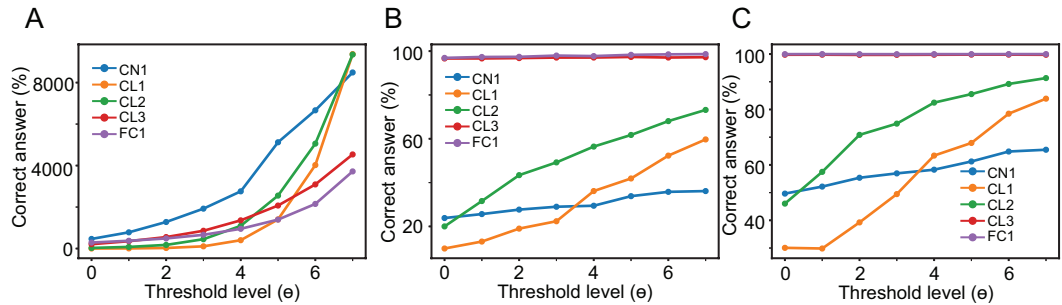}}
		\caption{Empirical evaluations using the ResNet trained for CIFAR 10. (A), The sizes of the library networks (i.e., the number of output nodes) depending on threshold values $\theta$ for novelty detection. Since HAPs' homogeneity varies significantly from one layer to another, a widely different set of threshold values is chosen for each layer. The actual values are listed in Table 1. (B), The fraction of correct predictions of the library networks using single best answers. (C), The same as (B), but the three best answers are used for predictions. }
		\label{Fig4}
	\end{center}
	
\end{figure}

We note that the diversity of HAPs in the ResNet strongly varies from one block to another, and thus widely different threshold values $\theta$ are necessary to describe them properly. Specifically, we use 8 sets of $\theta$s for CN1, CL1-3 and FC, respectively; see  Table \ref{table1} for the actual values. The library networks' sizes built for CN1,CL1-3 and FC1 are shown in Fig. \ref{Fig4}A. Figures \ref{Fig4}B and C show the accuracies of 5 library networks' predictions depending on the 8 sets of $\theta$s. Although the accuracies are lower than those of MNIST case (Fig. \ref{Fig2}), the library networks can still reliably predict the ResNet's decisions on test examples (Fig. \ref{Fig4}B and C), which indicates that even HAPs in the ResNet are forced to cluster during training. We also note that HAPs in CL1 (e.g., the first composite layer) are more homogeneous than those in any other blocks/layers in the ResNet (Fig. \ref{Fig4}A), which is distinct from the gradual increase in HAPs' homogeneity in the CNN (Fig. \ref{Fig2}A). This suggests that the main function of the RenNet's first block (CL1) is to transform features detected by CN1 into more homogeneous vectors and that the function of subsequent blocks/layers is to find optimal ways to map these highly homogeneous vectors (from the first block) into proper clusters to reflect the classes of input patterns.  

\subsection{Adversarial attacks reduce consistency among library networks' predictions}
The results above indicate that the library networks can predict DNNs' answers and that multiple library networks provide multiple predictions. In our experiments, we note that the library networks' predictions are largely consistent with one another and thus assume that the library networks’ consistency among their predictions (CPL, hereafter) is the result of training; if DNNs are well trained, all parts of the networks are consistent with one another. This leads us to speculate that CPL could decrease when inputs are drawn from `out-of-training' domains (i.e., novel domains). We address this hypothesis by introducing adversarially manipulated inputs and measuring CPL with correlations between prediction node outputs of the library networks (Eq. \ref{eq5}).
\begin{equation}\label{eq5}
CPL(k)=\frac{1}{N}\sum_{i > j} \frac{\vec{Prob^k_i} \boldsymbol{\cdot} \vec{Prob^k_j}} {\left\Vert\vec{Prob^k_i}\right\Vert \left\Vert\vec{Prob^k_j}\right\Vert}, 
\end{equation}
\begin{equation*}
\vec{Prob^k_i}=\frac{1}{\sum_l (\exp(P^k_l))} \left [ \exp(P^k_1),...,\exp(P^k_{10}) \right ], 
\end{equation*}
, where $P^k_l$ is likelihood (Eq. \ref{eq3}) estimated from library network built for RestNet's block $i$ or CNN's hidden layer $i$ elicited by $k$th input pattern. For both ResNets and CNNs, 20 maximally activated outputs of library networks are used to estimate  $P^k_l$, and $N$ is the number of all possible pairs of blocks/layers. By definition, we get a single CPL value for an input pattern. 

\begin{figure}[ht]
	
	\begin{center}
		\centerline{\includegraphics[width=5.5in]{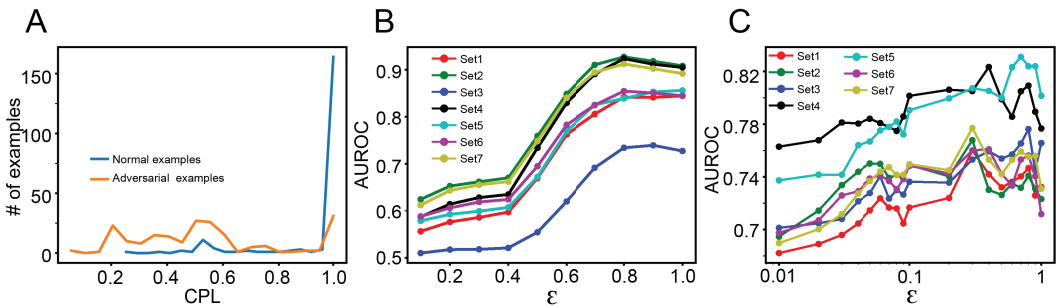}}
		\caption{Detecting adversarial attacks: (A), The histogram of comparing CPL values between normal and adversarial examples. (B), AUROC calculated using CPL values from the CNN trained for MNIST, depending on the $\epsilon$. The colors represent a set of threshold values chosen for the library networks; see Table \ref{table1} for the actual values. (C), The same as (B), but AUROC calculated from the ResNet.}
		\label{Fig5}
	\end{center}
	
\end{figure}

In our study, we use the routine `LinfPGDAttack' included in the `advertorch' \citep{ding2019advertorch}, which implements the projected gradient descent attack \citep {Madry2019}, to generate 200 adversarial images from MNIST and CIFAR 10 datasets, respectively. We fix the iteration number at 40 and step size at 0.01, while we test multiple perturbation sizes $\epsilon$. CPLs are estimated according to Eq. \ref{eq5} for 200 normal and adversarial images. To evaluate CPLs' dependence on a threshold value $\theta$ of each layer, we test 7 sets of threshold values (see Table \ref{table1}). Figure \ref{Fig5}A shows an example of CPL distributions calculated with normal and adversarial images of MNIST dataset. As shown in the figure, CPLs are distinct and well separated between normal and adversarial inputs. To quantify how well they can be separated by an ideal observer, we calculate the area of the receiver operating characteristic curve (AUROC); see \citep{wiki:xxx} for details. Figure \ref{Fig5}B shows the changes in AUROC depending on the degree of adversarial attacks  ($\epsilon$). The color codes represent the selected set of threshold values $\theta$s. We also test CPLs of the ResNet between normal and adversarial inputs. As shown in Fig. \ref{Fig5}C, the AUROC values are generally higher than 0.7, indicating that the library networks can also detect adversarial images of CIFAR-10. Based on these results, we propose that CPL can be used to detect adversarial attacks or more broadly the out-of-training examples (see below). 

\section{Discussion}

To probe DNNs' decision-making processes, we studied the functional links between HAPs and DNN's decisions using the library networks. Our empirical evaluations show that the library networks can reliably predict DNN’s answers, suggesting that they can identify (internal) neural codes crucial for their decisions. We further found 1) that the library networks could allow us to infer functions of hidden layers and 2) that they could detect adversarial attacks. Thus, we propose that the library networks can help us build more explainable and safer DNNs.  

While in line with the earlier studies \citep{Alain2016,Montavon2011,Liu2018,Raghu2017} focusing on the properties of hidden neurons to explore DNNs' operating principles, our study differs in that 1) it emphasizes the importance of similarities among HAPs in understanding DNNs' decision-making processes, 2) it proposes the library network and prediction nodes that can help us better understand DNNs' decision-making processes and 3) it proposes potential links between CPL (consistency of predictions of library networks) and the detection of adversarial attacks or examples from novel domains. Below we discuss the implications of our study in detail. 

\subsection{Transparency of  DNNs' decisions and library networks}
The library networks allow us to inspect how well the hidden layer activity is correlated with DNNs' answers. While the estimated correlations may not fully explain DNNs' operations, they do allow us to look into their decision-making processes. In our CNN experiments, the confusion matrices suggest that 1) CN1 is trained to detect necessary visual features, 2) CN2 is trained to amplify differences among the features, and 3) FC1 and 2 are trained to find ideal ways to utilize global features like locations to make decisions. With hidden layers' potential functions, we can make an effective flowchart of DNNs' decision processes.

Furthermore, we note that it is possible to use the library networks to estimate individual layers' performance and then selectively improve under-performing layers. If a library network suggests that two classes cannot be clearly separated from each other in one of the CN layers (or a functional block of DNNs assigned to feature-detection), we can conclude that more filters are needed for the layer (or blocks) and provide additional filters to improve its resolution power.  Alternatively, if a library network suggests that a FC layer performs poorly, we can retrain it while keeping the other parts intact or add more FC layers to the network to enhance DNNs' performance.  

\subsection{Detecting out-of-training examples via library networks}
DNNs can work properly only when test inputs and training examples are drawn from the same domain. More importantly, after being trained, DNNs implicitly assume that the current inputs originate from the domain (to which they had been exposed previously). For instance, once DNNs are trained with MNIST, they will always report numerical digits as the best answers even when the inputs are alphabet letters. These indiscriminate answers to `out-of-training' examples may result in critical errors under particular circumstances. Thus, DNNs need automatic systems to detect out-of-training examples to prevent catastrophic failures. We note that adversarially manipulated inputs can be considered as out-of-training examples. Together with our observation (\ref{Fig5})  that CPL is reduced when adversarially manipulated inputs are introduced, we propose that the library networks and CPL can be used to determine whether the inputs originate from the proper domains; when CPL values are too low, we should seek a second opinion (i.e., alternative DNNs or human opinion) or retrain the networks. 

\subsection{Future directions}
In this study, we maintain the structure of the library networks and learning algorithms for prediction nodes (Eq. \ref{eq2}) as simple as possible. Instead of refining them, we focused on addressing the potential functional links between hidden layer activity patterns and DNNs' decisions. Even though the library networks and learning algorithms for prediction nodes are not highly optimized, they can still successfully predict DNNs' answers on test examples, suggesting that HAPs are effectively clustered according to their classes during training. 

We believe that these functional clusters of HAPs can shed light on DNNs’ decision-making processes. Thus, in the future, we will extend the library networks to further study functional clusters of HAPs in two ways. First, we will test the capability of new library networks which will take a subset of hidden neurons (chosen sparsely and randomly) from the same hidden layers. This `sparse sampling' can reduce the library networks' sizes, which may be necessary for a large-scale dataset or highly complex network architectures, and we will study the predictive powers of the `sparse sampling library networks'. Second, we will construct library networks that sparsely sample  hidden neurons across layers to determine whether DNNs can develop internal concepts necessary for given tasks. If a library network detects a crucial set of hidden neurons across hidden layers, it suggests that DNNs can develop certain task-specific `concepts' and use them to perform tasks. We believe that these hypothetical concepts, if they exist, can help us better understand DNNs' operating principles and build truly explainable DL/DNNs.

\begin{table}[ht]
	\caption{We list the threshold ($\theta$) values that are used to build the library networks for the ResNet. The top rows show the ranges of $\theta$ tested for all 5 layers. The bottom rows show the $\theta$ chosen for calculating CPLs.}
	\begin{tabular}{|l|llllllll}
		\hline
		Level & 0    & 1    & 2    & 3    & 4                         & 5    & 6    & \multicolumn{1}{l|}{7}    \\ \hline
		CN1   & 0.18 & 0.2  & 0.22 & 0.24 & 0.26                      & 0.3  & 0.32 & \multicolumn{1}{l|}{0.34} \\
		CL1    & 0.64 & 0.66 & 0.68 & 0.7  & 0.72                      & 0.74 & 0.76 & \multicolumn{1}{l|}{0.78} \\
		CL2    & 0.48 & 0.5  & 0.52 & 0.54 & 0.56                      & 0.58 & 0.6  & \multicolumn{1}{l|}{0.62} \\
		CL3    & 0.5  & 0.52 & 0.54 & 0.56 & 0.58                      & 0.6  & 0.62 & \multicolumn{1}{l|}{0.64} \\
		FC    & 0.8  & 0.82 & 0.84 & 0.86 & 0.88                      & 0.9  & 0.92 & \multicolumn{1}{l|}{0.94} \\ \hline
		& CN1  & L1   & L2   & L3   & \multicolumn{1}{l|}{FC}   &      &      &                           \\ \cline{1-6}
		Set1 & 0.24 & 0.72 & 0.58 & 0.62 & \multicolumn{1}{l|}{0.94} &      &      &                           \\
		Set2 & 0.26 & 0.72 & 0.56 & 0.58 & \multicolumn{1}{l|}{0.88} &      &      &                           \\
		Set3 & 0.3  & 0.74 & 0.58 & 0.6  & \multicolumn{1}{l|}{0.9}  &      &      &                           \\
		Set4 & 0.32 & 0.76 & 0.6  & 0.62 & \multicolumn{1}{l|}{0.92} &      &      &                           \\
		Set5 & 0.34 & 0.78 & 0.62 & 0.64 & \multicolumn{1}{l|}{0.94} &      &      &                           \\
		Set6 & 0.26 & 0.72 & 0.6  & 0.62 & \multicolumn{1}{l|}{0.92} &      &      &                           \\
		Set7 & 0.26 & 0.72 & 0.62 & 0.64 & \multicolumn{1}{l|}{0.94} &      &      &                           \\ \cline{1-6}
	\end{tabular}
	\label{table1}
\end{table}


\acks{We thank the Allen Institute founder, Paul G. Allen, for his vision, encouragement and support. }


\newpage

\vskip 0.2in
\bibliography{ref}
\bibliographystyle{unsrt}
\end{document}